\RequirePackage{snapshot}
\documentclass{article} 
\usepackage{iclr2019_conference,times}

\usepackage{hyperref}
\usepackage{url}
\usepackage{subcaption}
\usepackage{mycustom}

\title{TherML:\\ { \Large The Thermodynamics of Machine Learning }}

\author{Alexander A.~Alemi \\
Google Research\\
1600 Amphitheatre Parkway \\
Mountain View, CA 94043 \\
\texttt{alemi@google.com} \\
\And
Ian Fischer \\
Google Research\\
1600 Amphitheatre Parkway \\
Mountain View, CA 94043 \\
\texttt{iansf@google.com}
}

\iclrfinalcopy 
\begin{document}

\maketitle

	\begin{abstract} 
	In this work we offer an information-theoretic framework for representation learning
	that connects with a wide class of existing objectives in machine learning.
	We develop a formal correspondence between this work and thermodynamics and discuss its
	implications.
\end{abstract}
}%

%

\section{Introduction}


Let $X,Y$ be some paired data, for example: a set of images $X$ and their labels $Y$. 
We imagine the data comes from some
\emph{true}, unknown data generating process $\Phi$\footnote{
	 Here we aim to invoke the same philosophy as in the introduction to \citet{watanabegreen}.}, from which we
 have drawn a \emph{training set} of $N$ pairs:
\begin{equation}
	\mathcal{T}_N \equiv (x^N, y^N) \equiv \{ x_1, y_1, x_2, y_2, \dots, x_N, y_N \} \sim \phi(x^N,y^N).
\end{equation}
We further imagine the process is \emph{exchangeable}\footnote{
	That is, we imagine the data satisfies De Finetti's theorem, for which infinite exchangeable processes usually
can be described by products of conditionally independent distributions, but don't want to worry too much about the
complicated details since there are subtle special cases~\citep{definetti}.}
and the data is conditionally independent given the governing
process $\Phi$:
\begin{equation}
	p(x^N, y^N | \phi) = \prod_i p(x_i|\phi)p(y_i|x_i,\phi).
\end{equation}
As machine learners, we believe that by
studying the training set, we should be able to infer or predict
new draws from the same data generating process.  Call a set of $M$ future
draws from the data generating process $\mathcal{T}'_M \equiv \{ X^M, Y^M \}$ the
\emph{test set}.

The \emph{predictive information}~\citep{predictive} 
is the \emph{mutual information} between the training set and a infinite test set,
 equivalently the amount of information the training set provides about the generative process itself:
\begin{equation}
	I_{\text{pred}}(\mathcal{T}_N) \equiv \lim_{M\to\infty} I(\mathcal{T}_N ; \mathcal{T}'_M) = I(\mathcal{T}_N; \Phi) = I(X^N, Y^N; \Phi).
\end{equation}
The predictive information measures the underlying complexity of the data generating process~\citep{still},
and is fundamentally limited and must grow sublinearly in the dataset size~\citep{predictive}.
%
Hence, the predictive information is a vanishing fraction of the total information in the training
set~\footnote{
 Here and throughout $H(A)$ is used to denote entropies $H(A) = - \sum_i p(A) \log p(A)$.
}:
\begin{equation}
	\lim_{N\to\infty} \frac{I_\text{pred}(\mathcal{T}_N)}{H(\mathcal{T}_N)} = 0
\end{equation}
A vanishing fraction of the information present in our training data is in any way
useful for future tasks.  A vanishing fraction of the information contained in the training
data is \emph{signal}, the rest is \emph{noise}. 
We claim the goal of learning
is to learn a \emph{representation} of data, both
locally and globally that captures the predictive information while
being maximally compressed: that separates the signal from the noise.

}%

\section{A Tale of Two Worlds}

We are primarily interested in
learning a stochastic local \emph{representation} of $X$, call it $Z$,
defined by some parametric distribution of our own design: 
$p(z_i|x_i,\theta)$ with its
own parameters $\theta$. 
A \emph{training procedure} is a process that assigns a distribution
$p(\theta|x^N, y^N)$
to the parameters conditioned on the observed dataset. In this way,
the parameters of our local parametric map are themselves a global
representation of the dataset.  With our augmentations,
the world now looks like the graphical model in Figure~\ref{fig:P}, denoted
World $P$: Some data generating process $\Phi$ generates a dataset $(X^N, Y^N)$ 
which we perform some learning algorithm on to get some parameters $p(\theta | x^N, y^N)$ 
which we can use to form a parametric local representation $p(z_i | x_i, \theta)$.

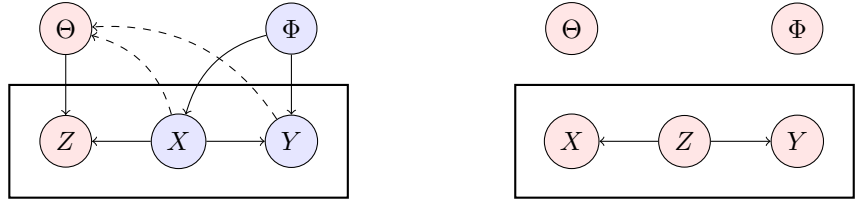
\begin{figure}[tbp]
	\centering
	\begin{subfigure}[t]{0.55\textwidth}
	\centering
	\begin{tikzpicture} 
\def\r{0.5 cm}
\node [draw, circle,fill=blue!10] (X) at (0, 0) {$X$};
\node [draw, circle,fill=blue!10] (Y) at (3*\r, 0) {$Y$};
\node [draw, circle,fill=red!10] (Z) at (-3*\r, 0) {$Z$};
\node [draw, circle,fill=red!10] (theta) at (-3*\r, 3*\r) {$\Theta$};
\node [draw, circle,fill=blue!10] (phi) at (3*\r, 3*\r) {$\Phi$};
\draw [->] (X) -- (Z);
\draw [->] (X) -- (Y);
\draw [->] (theta) -- (Z);
\draw [->] (phi) edge[bend right] (X);
\draw [->] (phi) -- (Y);
\draw [->] (X) edge[bend right, dashed] (theta);
\draw [->] (Y) edge[bend right, dashed] (theta);

\draw [thick] (-4.5*\r, -1.5*\r) rectangle (4.5*\r, 1.5*\r);

\end{tikzpicture}
	\caption{\label{fig:P}Graphical model for world $P$, 
		the real world augmented with a local
		and global representation.
		The dashed lines emphasize that $\theta$ only 
		depends on the first
		$N$ data points, the training set.
	Blue denotes nodes outside our control, while red nodes are under our direct control.
		}
	\end{subfigure}
	\quad
	\begin{subfigure}[t]{0.35\textwidth}
	\centering
	\begin{tikzpicture} 
\def\r{0.5 cm}
\node [draw, circle,fill=red!10] (Z) at (0, 0) {$Z$};
\node [draw, circle,fill=red!10] (Y) at (3*\r, 0) {$Y$};
\node [draw, circle,fill=red!10] (X) at (-3*\r, 0) {$X$};
\node [draw, circle,fill=red!10] (theta) at (-3*\r, 3*\r) {$\Theta$};
\node [draw, circle,fill=red!10] (phi) at (3*\r, 3*\r) {$\Phi$};
\draw [->] (Z) -- (X);
\draw [->] (Z) -- (Y);

\draw [thick] (-4.5*\r, -1.5*\r) rectangle (4.5*\r, 1.5*\r);

\end{tikzpicture}
	\caption{\label{fig:Q}Graphical model for world $Q$, the world we
	desire.  In this world, $Z$ acts as a latent variable for $X$ and $Y$ jointly.
	}
	\end{subfigure}
	\caption{\label{fig:graphical}Graphical models.
	}
\end{figure}

World $P$ is what we have.
It is not necessarily what we want.
What we \emph{have} to contend with is an unknown distribution of our data.
What we \emph{want} is a world that corresponds to the 
traditional modeling assumptions in which $Z$ acts
as a latent factor for $X$ and $Y$, rending them conditionally independent, leaving
 no correlations unexplained.
Similarly, we would prefer if we could easily marginalize out
the dependence on our universal ($\Phi$) and model specific ($\Theta$) parameters.
World $Q$ in Figure~\ref{fig:Q} is the world we \emph{want}~\footnote{
	We could consider different alternatives, deciding to relax some of the constraints we imposed
	in World $Q$, or generalizing World $P$ by letting the representation depend on $X$ and $Y$ jointly, 
	for instance.  What follows demonstrates a general sort of \emph{calculus} that we can invoke
	for any specified pair of graphical models. In particular~\Cref{sec:3d,sec:elwood,sec:discriminative}
	discuss alternatives.}.

We can measure the degree to which the real world aligns with our
desires by computing the minimum possible \emph{relative information}\footnote{
	Also known as the KL divergence.}
between our distribution $p$ and any distribution consistent with the 
conditional dependencies encoded in graphical model $Q$\footnote{
	Note that this is $\KL{p}{q^*}$ where $q^*$ is the 
	well known reverse-\emph{information projection} or \emph{moment projection}: 
	$q^* = \argmin_{q \in Q} \KL{p}{q}$~\citep{infoprojection}.
	}.
It can be shown~\citep{mib} that this quantity is given by the difference in
multi-informations between the two graphical models, as measured
in World $P$:
\begin{equation}
	\mathcal{J} \equiv \min_{q\in Q} \KL{p}{q} = I_P - I_Q .
\end{equation}

The \emph{multi-information}~\citep{multiinformation}
of a graphical model is the KL divergence between the joint distribution
and the product of all of the marginal distributions, which can be computed as a sum of mutual informations,
one for each node in the graph, between itself and its parents:
\begin{equation}
	I_G \equiv \left\langle \log \frac{p(g^N)}{ \prod_i p(g_i)} \right\rangle = \sum_i I(g_i; \operatorname{Pa}(g_i) )
\end{equation}


In our case:
\begin{equation}
	\mathcal{J} = I(\Theta; X^N,Y^N) + \sum_i\left[ I(X_i; \Phi) + I(Y_i ; X_i, \Phi) + I(Z_i; X_i, \Theta) - I(X_i; Z_i) - I(Y_i;Z_i) \right].
	\label{eqn:minkl}
\end{equation}
This minimal relative information has two terms outside our control and
we can take them to be constant, but which relate to the predictive information:
\begin{equation}
	\sum_i \left[ I(X_i; \Phi) + I(Y_i; X_i , \Phi) \right] \geq \sum_i I(Y_i; X_i) + I_{\text{pred}}(\mathcal{T}_N) .
\end{equation}
These terms measure the intrinsic complexity of our data. The remaining four terms are:
\begin{itemize}
	\item $I(X_i;Z_i)$ - which measures how much information our
		representation contains about the input ($X$). This should be
		maximized to ensure our local representation actually represents the
		input.
	\item $I(Y_i;Z_i)$ - which measures how much information our
		representation contains about our auxiliary data. This should
		be maximized as well to ensure that our local representation is predictive
		for the labels.
	\item $I(Z_i; X_i, \Theta)$ - which measures how much information
		the parameters and input determine about our representation.
		This should be minimized to ensure consistency between worlds, and
		ensure we learn compressed local representations.
		Notice that this is similar to, but distinct from the first
		term above.
		\begin{equation}
		I(Z_i; X_i, \Theta) = I(Z_i; X_i) + I(Z_i; \Theta | X_i)
			\label{eqn:ixzdiff}
		\end{equation}
		by the Chain Rule for mutual information~\footnote{Given this relationship,
		we could actually reduce the total number of functions we consider from 4 to 3, 
		as discussed in~\Cref{sec:3d}.}.
	\item $I(\Theta; X^N,Y^N)$ - which measures how much information
		we store about our training data in the parameters of our
		encoder. This should also be minimized to ensure we 
		learn compressed global representation, preveting overfitting.
\end{itemize}
These mutual informations are all intractable in general, 
since we cannot compute the necessary marginals in closed form,
given that we do not have access to the true data generating distribution.

\subsection{Functionals}
\label{sec:functionals}
Despite their intractability, we can compute variational bounds
on these mutual informations.

\subsubsection{Entropy}
\begin{equation}
	S \equiv \left\langle \log \frac{p(\theta| x^N,y^N)}{q(\theta)}
	\right\rangle_P \geq I(\Theta; X^N,Y^N)
	\label{eqn:entropy}
\end{equation}

The relative entropy in our parameters or just \emph{entropy} for short
measures the relative information between the distribution we assign
our parameters in World $P$ after learning from the data $(X^N,Y^N)$,
with respect to some data independent $q(\theta)$ \emph{prior}
on the parameters. This is an upper bound on the mutual information
between the data and our parameters and as such can measure our
risk of overfitting our parameters.

\subsubsection{Rate}

\begin{equation}
	R_i \equiv \left\langle \log \frac{p(z_i|x_i,\theta)}{q(z_i)} \right\rangle_P
	\geq I(Z_i;X_i,\Theta) 
	\label{eqn:rate}
\end{equation}

The \emph{rate} measures
the complexity of our representation.  It is the
relative information of a sample specific representation
$z_i \sim p(z|x_i,\theta)$ with respect to our
variational marginal $q(z)$.  It measures how many bits
we actually encode about each sample, and can measure how our
risk of overfitting our representation. We use $R \equiv \sum_i R_i$.

\subsubsection{Classification Error}
\begin{equation}
	C_i \equiv -\left\langle \log q(y_i|z_i) \right\rangle_P
	\geq H(Y_i) - I(Y_i;Z_i) = H(Y_i|Z_i) 
	\label{eqn:classificationerror}
\end{equation}

The \emph{classification error} measures the conditional entropy of $Y$
left after conditioning on $Z$. It is a measure of how much
information about $Y$ is left unspecified in our representation.
This functional measures our supervised learning performance.
We use $C \equiv \sum_i C_i$.

\subsubsection{Distortion}
\begin{equation}
	D_i \equiv - \left\langle \log q(x_i|z_i) \right\rangle_P \geq H(X_i) - I(X_i;Z_i)
= H(X_i|Z_i) 
	\label{eqn:distortion}
\end{equation}

The \emph{distortion} measures the conditional entropy of $X$
left after conditioning on $Z$.  It is a measure of how much
information about $X$ is left unspecified in our
representation. This functional measures our unsupervised learning
performance.
We use $D \equiv \sum_i D_i$.

\subsection{Geometry}

The distributions $p(z|x,\theta), p(\theta |
x^N,y^N), q(z) , q(x|z), q(y|z)$ can be chosen arbitrarily.
Once chosen, the \emph{functionals} $R, C, D, S$ take on well
described values.  The choice of the five distributional families
specifies a single point in a four-dimensional space.

Importantly, the sum of these functionals
is a variational upper bound (up to an additive constant)
for the minimum possible relative information between worlds~(\Cref{sec:upperboundproof}):
\begin{equation}
	S + R + C + D \geq \mathcal{J} + \sum_i H(X_i,Y_i | \Phi) 
	\label{eqn:upperbound}
\end{equation}

Besides just the upper bound, we can consider the full space of \emph{feasible} points.
Notice that $S$ and $R$ are both themselves upper bounds on mutual
informations, and so must be positive semi-definite.
If our data is discrete, or if we have discretized
it~\footnote{More generally, if we choose some measure $m(x),m(y)$ on
both $X$ and $Y$, we can define $D$ and $C$ in terms
of that measure e.g.
$D \equiv -\left\langle \log \frac{q(x|z)}{m(x)} \right\rangle_P \geq H_m(X) - I(X;Z) = H_m(X|Z)$
	},
$D$ and $C$ which are both upper bounds on conditional entropies, must be
positive as well. Along with \Cref{eqn:upperbound}, given that $\sum_i H(X_i, Y_i | \Phi)$ is
a positive constant outside our control, the space of possible $(R, C, D, S)$ values
is at least restricted to be points in the positive orthant with some minimum possible Manhattan distance to the origin:
\begin{equation}
	S + R + C + D \geq \sum_i H(X_i, Y_i | \Phi) \qquad R \geq 0 \quad S \geq 0 \quad D \geq 0 \quad C \geq 0
	\label{eqn:polytope}
\end{equation}
Even in the infinite model family limit, data-processing inequalities on mutual information terms
all defined in a set of variables that satisfy some nontrivial conditional dependencies ensure that
there are regions in this functional space that are wholly out of reach.  The surface of the feasible
region maps an optimal frontier, optimal in the degree to which it minimizes mismatch between our two worlds
subject to constraints on the relative magnitudes of the individual terms.  
This convex polytope has edges, faces and corners
that are identifiable as the optimal solutions for well known objectives.

This story is a generalization of the story presented in \citet{brokenelbo}, which can be considered
a two-dimensional projection of this larger space (onto $R, D$).  Within our larger framework we can derive
more specific bounds between subsets of the functionals. For instance:
\begin{equation}
	R_i + D_i \geq H(X_i) + I(Z_i; \Theta | X_i) .
\end{equation}
This mirrors the bound given in \citet{brokenelbo} where $R+D \geq H(X)$,
which is still true given that all conditional mutual informations are positive
semi-definite $(H(X) + I(Z;\Theta|X) \geq H(X))$,
but here we obtain a tighter pointwise bound that has a term
measuring how much information about our encoding is revealed
by the parameters after conditioning on the input itself.  This term
$I(Z_i;\Theta|X_i)$  captures the degree to which our local representation is
overly sensitive to the particular parameter settings~\footnote{In~\Cref{sec:3d} we consider
	taking this bound seriously to limit the space only only
	three functionals, $S$, $C$ and $V \geq I(Z_i;\Theta|X_i)$}\footnote{
		This could help explain the observation that often times putting additional modeling power
		on the prior rather than the encoder can give improvements in ELBO~\citep{vlossy}.
		}.

}%

	\subsection{Generalization}

We can evaluate how much information our representations capture
about the true data generating process.
For instance, $I(Z_i; \Phi)$ which measures how much information
about the true data generating procedure our local representations 
capture.  Notice that given
the conditional dependencies in world $P$, we have the following Markov
chain:
\begin{equation}
	\Phi \rightarrow (X_i, Y_i, \Theta) \rightarrow Z_i
\end{equation}
and so by the Data Processing Inequality~\citep{coverthomas}:
\begin{equation}
	I(Z_i;\Phi) \leq I(Z_i;\Theta,X_i,Y_i) 
	= I(Z_i;X_i,\Theta) + \cancel{I(Z_i; Y_i | X_i,\Theta)} \leq R_i.
\end{equation}

The per-instance rate $R_i$ forms an upper bound on the mutual information between
our encoding $Z_i$ and the \emph{true} governing parameters of our data $\Phi$.
Similarly, we can establish that:
\begin{equation}
	\Phi \rightarrow (X^N, Y^N) \rightarrow \Theta \implies I(\Theta; \Phi) \leq I(\Theta; X^N,Y^N) \leq S .
\end{equation}
$S$ upper bounds the amount of information our encoder's parameters $\Theta$,
the global representation of the dataset 
can contain about the true process $\Phi$. At the same time:
\begin{equation}
	I(\Theta; \Phi) \leq I(X^N,Y^N ; \Phi) \leq \sum_i I(X_i, Y_i; \Phi),
\end{equation}
which sets a natural upper limit for the maximum $S$ that might be useful.


}%

\section{Optimal Frontier}
\label{sec:frontier}

As in~\citet{brokenelbo},
under mild assumptions about the variational distributional families, 
it can be argued that
the surface is monotonic in all of its arguments. The optimal surface in the
infinite family limit can be characterized as a convex polytope~(\Cref{eqn:polytope}).
In practice we will be in the realistic setting corresponding to finite parametric families such as
neural network approximators.
 We then expect that there is an irrevocable gap that opens up in the variational
bounds.  
Any failure of the distributional families to model the correct
corresponding marginal in $P$ means that the space of all \emph{realizable}
$R,C,D,S$ values will be some convex relaxation of the optimal \emph{feasible} surface.
This surface will be described some function $f(R,C,D,S) = 0$,
which means we can identify points on the surface as a function of one
functional with respect to the others (e.g. $R = R(C,D,S)$).
Finding points on this surface equates to solving a constrained optimization
problem, e.g.
\begin{equation}
	\min_{q(z)q(x|z)q(y|z)p(z|x,\theta)p(\theta|\{x,y\})} R 
	\text{ such that } D = D_0, S = S_0, C = C_0.
\end{equation}
Equivalently, we could solve the unconstrained Lagrange multipliers problem:
\begin{equation}
	\min_{q(z)q(x|z)q(y|z)p(z|x,\theta)p(\theta|\{x,y\})} R + \delta D + \gamma C + \sigma S.
	\label{eqn:objective}
\end{equation}
Here $\delta, \gamma, \sigma$ are Lagrange multipliers that impose the
constraints. They each correspond to the partial derivative of the rate
at the solution with respect to their corresponding functional, keeping the
others fixed.

Notice that this single objective encompasses a wide range of existing
techniques.
\begin{itemize}
\item If we retain $C$ alone, we are doing traditional supervised
	learning
	and our network will learn to be deterministic in its activations and
	parameters.
\item If $\delta = 0$ we no longer require a variational reconstruction network
	$q(x|z)$, and are doing some form of supervised learning generally.
\item If $\delta = 0, \sigma = 0$ we exactly recover the Variational
	Information Bottleneck (VIB) objective of
	\citet{vib} (where $\beta = 1/\gamma$),
	a form of stochastically regularized supervised learning that
	imposes a bottleneck on how much information our representation
	can retain about the input, while simultaneously
	maximizing the amount of information the representation contains
	about the target.
\item If $\delta = 0$ and $\sigma, \gamma \to \infty$ but in
	such a way as to keep the ratio fixed $\beta \equiv \sigma/\gamma$
	(that is if we drop the $R$ term and only keep $C + \beta S$ as our
	objective) we recover the Information Bottleneck Lagrangian 
		loss of \citet{emergence}, presented
	as an alternative way to do Information Bottleneck~\citep{ib} but
	being stochastic on the parameters rather than the activations as in
		VIB.
\item As a special case, if our objective is set to $C + S$ ($\delta = 0,
	\sigma,\gamma \to \infty, \sigma/\gamma \to 1$), we obtain
	the objective for a Bayesian neural network, ala \citet{bbb}.

\item If we retain only $D$, we are training a stochastic autoencoder.
\item If $\sigma = 0, \gamma = 0, \delta = 1$ the objective is
equivalent to the ELBO used to train a VAE~\citep{vae}.
\item If $\sigma = 0, \gamma = 0$ more generally, the objective is equivalent
	to a $\beta$-VAE~\citep{betavae} where $\beta = 1/\delta$.
\item If $\gamma = 0$ all terms involving the auxiliary data $Y$ drop out
	and we are doing some form of unsupervised learning without any
	variational classifier $q(y|z)$. The presence
	of the $S$ term makes this more general than a usual $\beta$-VAE
	and should offer better generalization properties and control
	of overfitting by bottle-necking how much information we allow the
	parameters of our encoder
	to extract from the training data.
\item $\sigma = 0, \gamma = \alpha, \delta = 1$ recovers the semi-supervised
	objective of \citet{semi}.
\item In its most general form, in common parlance the full objective might be 
	described as a temperature-regulated Bayesian 
	semi-supervised $\beta$-VAE, or a
	Variational Information Bottleneck Lagrangian Autoencoder (VIBLA).
\end{itemize}
Examples of all of these objectives behavior on a simple toy model
is shown in Appendix~\ref{sec:experiments}.

Notice that all of these previous approaches describe low dimensional
sub-surfaces of the optimal three dimensional frontier.
These approaches were all interested in
different domains, some were focused on supervised prediction accuracy,
others on learning a generative model. Depending on your specific problem, and
downstream tasks, different points on the optimal frontier will be desirable.
However, instead of choosing a single point on the frontier,
we can now
explore a region on the surface to see
what class of solutions are possible within the modeling choices. By simply
adjusting the three control parameters $\delta, \gamma, \sigma$, we can smoothly
move across the entire frontier and smoothly interpolate between all of these
objectives and beyond.

\subsection{Optimization}

So far we've considered explicit forms of the objective in terms of the
four functionals.  For $S$ this would require some kind
of tractable approximation to the posterior over the parameters of our
encoding distribution\footnote{
	As in \citet{bbb, emergence}}.
Alternatively, we can formally describe the exact
solution to our minimization problem:
\begin{equation}
	\min S \text{ s.t. } R=R_0, C=C_0, D=D_0.
\end{equation}
Recall that $S$ measures the relative entropy of our parameter distribution
with respect to the $q(\theta)$ \emph{prior}. As such, the solution
that minimizes the relative entropy subject to some constraints is a
generalized Boltzmann distribution~\citep{jaynes}:
\begin{equation}
	p^*(\theta | \{x,y\}) =
	\frac{q(\theta)}{\mathcal{Z}} e^{-(R + \delta D + \gamma C)/\sigma}.
	\label{eqn:posterior}
\end{equation}
Here $\mathcal{Z}$ is the
\emph{partition function}, the normalization constant
for the distribution
\begin{equation}
	\mathcal{Z} = \int d\theta\, q(\theta)\, e^{-( R + \delta D + \gamma C)/\sigma}
	\label{eqn:partition}
\end{equation}

This suggests an alternative method for finding points on the optimal
frontier.  We could turn the unconstrained Lagrange optimization problem
that required some explicit choice of tractable posterior distribution
over parameters into a sampling problem for a richer implicit distribution.

A naive way to draw samples from this posterior would be to use Stochastic
Gradient Langevin Dynamics or its cousins~\citep{sgld,sghmc,sgr} which,
in practice, would look like ordinary stochastic gradient descent (or its
cousins like momentum) for the objective $R + \delta D + \gamma C$,
with injected noise.  By choosing the magnitude of the noise relative
to the learning rate, the effective temperature $\sigma$ can be
controlled.

There is increasing evidence that the stochastic part of stochastic gradient
descent itself is enough to turn SGD less into an optimization procedure
and more into an approximate
posterior sampler~\citep{sgdasbayes,bayessgd,emergence,energyentropy,soattosgd},
where hyperparameters such as the learning
rate and batch size set the effective temperature.
If ordinary
stochastic gradient descent is doing something more akin to sampling from a
posterior and less like optimizing to some minimum, it would help explain
improved performance through ensemble averages of different points along
trajectories~\citep{snapshot}.

When viewed in this light, Equation~\ref{eqn:posterior} describes the
optimal posterior for the parameters so as to ensure the minimal divergence
between worlds $P$ and $Q$. $q(\theta)$ plays the role of the \emph{prior}
over parameters, but our overall objective is minimized when
\begin{equation}
	q(\theta) = p(\theta) = \langle p(\theta|x^N,y^N) \rangle_{p(x^N,y^N)}.
\end{equation}
That is, when our \emph{prior} is the marginal of the posteriors over all possible
datasets drawn from the true distribution. A fair draw from this marginal is
 to take a sample from the posterior obtained on a different but related
dataset.  Insomuch as ordinary SGD training is an approximate method for
drawing a posterior sample, the common practice of fine-tuning a pretrained
network on a related dataset is using a sample from the optimal \emph{prior}
as our initial parameters.
The fact that fine-tuning approximates use of an optimal \emph{prior}
presumably helps explain its broad success.  

If we identify our true goal not as optimizing some objective but instead
directly sampling from Equation~\ref{eqn:posterior}, we can consider
alternative approaches to define our learning dynamics, such as \emph{parallel
tempering} or \emph{population annealing}~\citep{poppar}.  Alternatively, we
could, instead of adopting variational bounds on the mutual informations,
consider other mutual information bounds such as those in \citet{mine,cpc}.
Perhaps our priors can be fit, providing we form estimates of the expectation
over datasets (e.g. bootstrapping or jackknifing our
dataset~\citep{bootstrap}).

}%

\section{Thermodynamics}

So far we have described a framework for learning that
involves finding points
that lie on the surface of a convex three-dimensional surface in terms of four
functional coordinates $R,C,D,S$. Interestingly, this is all that is required
to establish a formal connection to thermodynamics, which
similarly is little more than 
the study of exact differentials~\citep{sethna,finn}.

Whereas previous approaches connecting thermodynamics and
learning~\citep{thermoinfo,costbenefitdata,thermoprediction} have focused on
describing the thermodynamics and statistical mechanics of physical
realizations of learning systems (i.e. the heat bath in these papers is a
physical heat bath at finite temperature), in this work we make a formal
analogy to the structure of the theory of thermodynamics, without any physical
content.

\subsection{First Law of Learning}

The optimal frontier creates an equivalence
class of states, being the set of all states that minimize as much as possible
the distortion introduced in projecting world $P$ onto a set of distributions
that respect the conditions in $Q$. The surface satisfies some
equation $f(R,C,D,S) = 0$ which we can use to describe any one
of these functionals in terms of the rest, e.g. $R = R(C,D,S)$.  This function
is entire, and so we can equate partial derivatives of the function with
differentials of the functionals\footnote{$\pdd{X}{Y}{Z}$ denotes the 
partial derivative of $X$ with respect to $Y$ holding $Z$ constant.}:
\begin{equation}
	dR = \pdd{R}{C}{D,S} dC + \pdd{R}{D}{C,S} dD + \pdd{R}{S}{C,D} dS.
\end{equation}
Since the function is smooth and convex, instead of identifying the surface of 
optimal rates in terms of the functionals $C,D,S$, we could just as well
describe the surface in terms of the partial derivatives by applying a Legendre
transformation. We will name the partial derivatives:
\begin{equation}
	\gamma \equiv -\pdd{R}{C}{D,S} \qquad
	\delta \equiv -\pdd{R}{D}{C,S} \qquad
	\sigma \equiv -\pdd{R}{S}{C,D} .
\end{equation}
These measure the exchange rate
for turning rate into reduced distortion, reduced classification error, or 
increased entropy, respectively.

The functionals $R,C,D,S$ are analogous to extensive thermodynamic variables
such as volume, entropy, particle number, magnetic field, charge, 
surface area, length and energy which 
grow as the system
grows, while the named partial derivatives $\gamma,\delta,\sigma$ are analogous
to the intensive, generalized forces in thermodynamics corresponding to their
paired state variable, such as pressure, temperature, chemical potential,
magnetization, electromotive force, surface tension, elastic force, etc.
Just as in thermodynamics, the \emph{extensive} functionals are defined for any
state, while the \emph{intensive} partial derivatives are only well defined
for \emph{equilibrium states}, which in our language are the states 
lying on the optimal surface~\footnote{For more discussion of equilibrium states,
and how they connect with more intuitive notions of equilibrium, see \Cref{sec:zeroth}}.

Recasting our total differential:
\begin{equation}
	dR = -\gamma dC - \delta dD - \sigma dS,
	\label{eqn:firstlaw}
\end{equation}
we create a law analogous to the \emph{First Law of Thermodynamics}. In
thermodynamics the First Law is often taken to be a statement about the 
conservation of energy, and by analogy here we could think about this 
\emph{law} as a statement about the conservation of information. Granted,
the actual content of the law is fairly vacuous, equivalent only to the 
statement that there exists a scalar function $R = R(C,D,S)$ defining our
surface and its partial derivatives.

\subsection{Maxwell Relations and Thermodynamic Potentials}

Requiring that Equation~\ref{eqn:firstlaw} be an exact differential
has mathematically trivial but intuitively non-obvious implications that relate
various partial derivatives of the system to one another, akin to the
\emph{Maxwell Relations} in thermodynamics. For example, requiring that
mixed second partial derivatives are symmetric establishes that:
\begin{equation}
	\pddd{R}{D}{C}{} = \pddd{R}{C}{D}{} \implies 
	\pdd{\delta}{C}{D} = \pdd{\gamma}{D}{C}.
\end{equation}
This equates the result of two very different experiments.  In the experiment
encoded in the partial derivative on the left, one would measure
the change in the derivative of the $R-D$ curve ($\delta$) as a function
of the classification error ($C$) at fixed distortion ($D$).  On the right
one would measure the change in the derivative of the $R-C$ curve ($\gamma$)
as a function of the distortion ($D$) at fixed classification error ($C$).
As different as these scenarios appear, they are mathematically equivalent.
A full set of Maxwell relations can be found in \Cref{sec:maxwell}.

We can additionally take and name higher order partial derivatives, analogous
to the susceptibilities of thermodynamics like bulk modulus, the thermal
expansion coefficient, or heat capacities. 
For instance, we can define the analog of
heat capacity for our system, a sort of rate capacity at constant distortion:
\begin{equation}
	K_D \equiv \pdd{R}{\sigma}{D}.
	\label{eqn:capacity}
\end{equation}
Just as in thermodynamics, these susceptibilities may offer useful ways to
characterize and quantify the systematic differences between model families.
Perhaps general scaling laws can be found between susceptibilities
and network widths, or depths, or number of parameters or dataset size.
Divergences or discontinuities in the susceptibilities are the hallmark of
phase transitions in physical systems, and it is reasonable to expect to see 
similar phenomenon for certain models.

A great deal of first, second and third order partial derivatives in
thermodynamics are given unique names. This is because the quantities are
particularly useful for comparing different physical systems.  We expect
a subset of the first, second and higher order partial derivatives of the 
base functionals will prove similarly useful for comparing, quantifying, and
understanding differences between modeling choices.

\subsection{Second Law of Learning?}

Even when doing deterministic training,
training is non-invertible~\citep{reversible}, and
we need to contend with and track the
entropy ($S$) term. 
We set the parameters of our networks initially with a fair
draw from some prior distribution $q(\theta)$. The training procedure
acts as a Markov process on the distribution of parameters, transforming
it from the prior distribution into some modified distribution, the
posterior $p(\theta|x^N,y^N)$.
Optimization is a many-to-one function,
that in the ideal limiting case,
 maps all possible initializations to a single global optimum.  In this
limiting case $S$ would be divergent, and there is nothing to prevent
us from memorizing the training set.  

The Second Law of Thermodynamics states that the entropy of an isolated system
tends to increase.  All systems tend to disorder, and this places limits
on the maximum possible efficiency of heat engines.

Formally, there are many statements akin to the Second Law of Thermodynamics
that can be made about Markov chains generally~\citep{coverthomas}.
The central one is that for any for any two distributions $p_n, q_n$ 
both evolving
according to the same Markov process ($n$ marks the time step),
the relative entropy $\KL{p_n}{q_n}$ is monotonically decreasing with time.
This establishes that for a stationary Markov chain, the relative entropy
to the stationary state $\KL{p_n}{p_\infty}$ monotonically
decreases~\footnote{For discrete state Markov chains, this implies that if the stationary
distribution is uniform, the entropy of the distribution $H(p_n)$ is strictly
increasing.}.

In our language, we can make strong statements about dynamics that target
points on the optimal frontier, or dynamics that implement a relaxation
towards equilibrium. 
There is a fundamental distinction between states that live on the frontier
and those off of it, analogous to the distinction between equilibrium and 
non-equilibrium states in thermodynamics.  

Any equilibrium distribution can be expressed in the form~\Cref{eqn:posterior} and
identified by its partial derivatives $\gamma, \delta, \sigma$.  
If name the objective in~\Cref{eqn:objective}:
\begin{equation}
	J(\gamma, \delta, \sigma) \equiv R + \delta D + \gamma C + \sigma S,
\end{equation}
The value this objective takes for any equilibrium distribution can be shown
to be given by the log partition function~(\Cref{eqn:partition}):
\begin{equation}
	\min J(\gamma, \delta, \sigma) = -\sigma \log \mathcal{Z}(\gamma, \delta, \sigma)
\end{equation}
and the KL divergence between any distribution over parameters $p(\theta)$ and an equilibrium
distribution is:
\begin{align}
	\KL{p(\theta)}{p^*(\theta;\gamma,\delta,\sigma)} =  \Delta J  / \sigma \\
	\Delta J \equiv J^{\textrm{noneq}}(p; \gamma,\delta,\sigma) - J(\gamma, \delta, \sigma)
\end{align}
Where $J^{\textrm{noneq}}$ is the non-equilibrium objective:
\begin{equation}
	J^{\textrm{noneq}}(p; \gamma, \delta,\sigma) = \left\langle R + \delta D + \gamma C + \sigma S \right\rangle_{p(\theta)}.
\end{equation}

For a stationary Markov process whose stationary distribution is an equilibrium
distribution the KL divergence to the stationary distribution must
monotonically decrease each step.  This means the $\Delta J / \sigma$ must
decrease monotonically, that is our objective $J$ must decrease monotonically:
\begin{equation}
	J_{t=0} \geq J_{t} \geq J_{t+1} \geq J_{t=\infty}.
\end{equation}
Furthermore, if we use $q(\theta)$ as our prior over parameters, we know:
\begin{align}
	J_{t=0} &= \left\langle R + \delta D + \gamma C \right\rangle_{q(\theta)}  \\
	J_{t=\infty} &= -\sigma \log Z.
\end{align}

}%

	\section{Conclusion}

We have formalized representation learning as 
the process of
minimizing the distortion introduced
when we project the real world (World $P$) onto the world we desire 
(World $Q$). The projection is naturally described by a set of four functionals
which variationally bound relevant mutual informations in the real world.
Relations between the functionals describe an optimal three-dimensional surface
in a four dimensional space of \emph{optimal} states. 
A single learning objective targeting points on this optimal surface can express
a wide array of existing learning objectives spanning from unsupervised
learning to supervised learning and everywhere in between.
The geometry of the
optimal frontier suggests a wide array of identities involving the functionals
and their partial derivatives.  This offers a direct analogy to thermodynamics
independent of any physical content. By analogy to thermodynamics,
we can begin to develop new quantitative measures and relationships amongst
properties of our models that we believe will offer a new class of theoretical
understanding of learning behavior.
}%

	\subsubsection*{Acknowledgements}

The authors would like to Jascha Sohl-Dickstein, 
Mallory Alemi, Rif A. Saurous, Ali Rahimi, Kevin Murphy,
Ben Poole, Danilo Rezende and Matt Hoffman for helpful discussions
and feedback on the draft.

}%
	
\bibliography{bib}
\bibliographystyle{iclr2019_conference}

\clearpage
\appendix

	\section{Reconstruction Free Formulation}
\label{sec:3d}

We can utilize the Chain Rule of Mutual Information~(\Cref{eqn:ixzdiff}):
\begin{equation}
	I(Z_i; X_i, \Theta) = I(Z_i; X_i) + I(Z_i; \Theta | X_i),
\end{equation}
to simplify our expression for the minimum possible KL between worlds~(\Cref{eqn:minkl}),
and consider a reduced set of functionals~(compare to \Cref{sec:functionals}):
\begin{itemize}
\item $C_i \equiv -\left\langle \log q(y_i|z_i) \right\rangle_P
	\geq H(Y_i) - I(Y_i;Z_i) = H(Y_i|Z_i) $

The \emph{classification error}, as before.

\item $S \equiv \left\langle \log \frac{p(\theta| \{x,y\})}{q(\theta)}
	\right\rangle_P \geq I(\Theta; \{X,Y\})$

The \emph{entropy} as before.

\item $V_i \equiv \left\langle \log \frac{p(z_i | x_i, \theta) }{q(z_i|x_i)}\right\rangle_P \geq I(Z_i; \Theta | X_i)$

The \emph{volume} of the representation (for lack of a better term), which measures the mutual
information between our representation $Z$ and the parameters $\Theta$, conditioned
on the input $X$.  That is, this functional bounds how much of the information
in our representation can come from the learning algorithm, independent of the actual input.
\end{itemize}

\begin{figure}[htbp]
	\centering
	\begin{tikzpicture} 
\def\r{0.5 cm}
\node [draw, circle] (Z) at (0, 0) {$Z$};
\node [draw, circle] (Y) at (3*\r, 0) {$Y$};
\node [draw, circle] (X) at (-3*\r, 0) {$X$};
\node [draw, circle] (theta) at (-3*\r, 3*\r) {$\Theta$};
\node [draw, circle] (phi) at (3*\r, 3*\r) {$\Phi$};
\draw [<-] (Z) -- (X);
\draw [->] (Z) -- (Y);

\draw [thick] (-4.5*\r, -1.5*\r) rectangle (4.5*\r, 1.5*\r);

\end{tikzpicture}
	\caption{\label{fig:Q2}Modified graphical model for world $Q$, instead of \Cref{fig:Q},
	the world we desire which satisfies the joint density in
	Equation~\ref{eqn:Q2}. Notice that this graphical model encodes all of the same conditional
	independencies as the original.
	}
\end{figure}
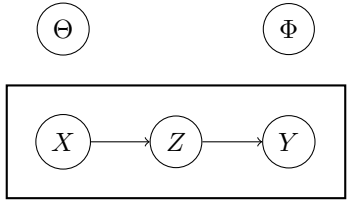

In principle, these three functionals still fully characterize the distortion introduced in
our information projection.
Notice that this new functional requires the variational approximation $q(z_i|x_i)$, a variational
approximation to the marginal over our parameter distribution. Notice also that we no longer
require a variational approximation to $p(x_i|z_i)$.  That is, in this formulation we no
longer require any form of decoder, or synthesis in our original data space $X$. 
While equivalent in its information projection, this more naturally corresponds to the 
model of our desired world $Q$:
\begin{equation}
	q(x,y,\phi,z,\theta) = q(\phi)q(\theta)\prod_i q(z_i|x_i)q(y_i|z_i),
	\label{eqn:Q2}
\end{equation}
depicted below in~\Cref{fig:Q2}.  Here we desire, not the joint
generative model $X \leftarrow Z \rightarrow Y$, but the predictive model $X \rightarrow Z \rightarrow Y$.

In this case we have:
\begin{equation}
	C + S + V \geq \mathcal{J}  + \sum_i \left[ H(Y_i X_i | \Phi) - H(X_i) \right].
\end{equation}

We can imagine tracing out this, now three dimensional, frontier that still explores a space consistent
with our original graphical model, but wherein we no longer have to do any form of direct variational synthesis.
}%

	\section{Bayesian Inference}
\label{sec:elwood}

\begin{figure}[htbp]
	\centering
	\begin{subfigure}[t]{0.40\textwidth}
	\centering
	\begin{tikzpicture} 
\def\r{0.5 cm}
\node [draw, circle] (X) at (0, 0) {$X$};
\node [draw, circle] (Y) at (3*\r, 0) {$\Theta$};
\node [draw, circle] (Z) at (-3*\r, 0) {$\Phi$};

\draw [->] (Z) -- (X);
\draw [dashed,->] (X) -- (Y);

\draw [thick] (-1.5*\r, -1.5*\r) rectangle (1.5*\r, 1.5*\r);

\end{tikzpicture}
	\caption{\label{fig:P}Graphical model for world $P$, 
	depicting Bayesian inference as learning a single global
		representation of data.  }
	\end{subfigure}
	\quad
	\begin{subfigure}[t]{0.40\textwidth}
	\centering
	\begin{tikzpicture} 
\def\r{0.5 cm}
\node [draw, circle] (X) at (0, 0) {$X$};
\node [draw, circle] (Y) at (3*\r, 0) {$\Theta$};
\node [draw, circle] (Z) at (-3*\r, 0) {$\Phi$};

\draw [->] (Y) -- (X);

\draw [thick] (-1.5*\r, -1.5*\r) rectangle (1.5*\r, 1.5*\r);

\end{tikzpicture}
	\caption{\label{fig:Q}Graphical model for world $Q$, the world we
	desire, the usual generative model of Bayesian inference.
	}
	\end{subfigure}
	\caption{\label{fig:bayes}Graphical models for standard Bayesian inference.}
\end{figure}
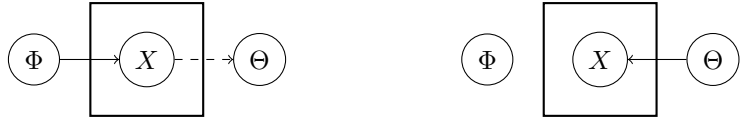

Just as in \ref{sec:3d} we can consider alternative graphical models for World
P.  In particular, we can consider a simplified scenario depicted in
\Cref{fig:bayes} corresponding to the usual situation in Bayesian inference.
Here we have just data, generated by some process and we form a single global
representation of the dataset. The world we desire, World Q, corresponds to the
usual Bayesian modeling assumption, whereby our own global representation
generates the data conditionally independently.

For these sets of graphical models, we have the following information projection:
\begin{equation}
	\mathcal{J_\text{bayes}} = \min_{q \in Q} \KL{p}{q} = I_P - I_Q = \sum_i I(X_i; \Phi) + I(\Theta; X^n) - \sum_i I(X_i; \Theta)
\end{equation}

And we can derive the simple variational bounds:

\begin{equation}
	S \equiv \left\langle \log \frac{p(\theta|X^n)}{q(\theta)}  \right\rangle \geq I(\Theta; X^n)
\end{equation}
This \emph{entropy} gives an upper bound on the mutual information between our parameters and the dataset,
it requires a variational approximation to the true marginal of the posterior $p(\theta | X^N)$ over datasets: $q(\theta)$,
a \emph{prior}.

\begin{equation}
	U_i \equiv - \left\langle \log q(x_i|\theta) \right\rangle \geq H(X_i | \Theta)
\end{equation}
The \emph{energy} gives an upper bound on the conditional entropy of our data given our parameters, it is powered
by a variational approximation to the factored inverse of our global representation, the \emph{likelihood} in ordinary 
parlance.

Our optimal frontier is set by those conditions above as well as:~\footnote{$U \equiv \sum_i U_i$}:
\begin{equation}
	U + S \geq \mathcal{J_{\text{bayes}}} + \sum_i H(X_i|\Phi)
\end{equation}

Just as in our earlier paper~\citep{brokenelbo} we could trace out the frontier by doing the constrained
optimization problem:
\begin{equation}
	\min S + \beta U 
\end{equation}

The formal solution to this optimization problem takes the form:
\begin{equation}
	\log p(\theta | x^N) = \log q(\theta) + \beta \sum_i \log q(x_i | \theta) - \log \mathcal{Z}.
\end{equation}
Where $\mathcal{Z}$ is the partition function:
\begin{equation}
	\mathcal{Z} = \int d\theta \, q(\theta) e^{\beta \sum_i \log q(x_i|\theta)}
\end{equation}

This is the ordinary temperature regulated~\citep{watanabegrey} Bayesian posterior:
\begin{equation}
	p(\theta | x^N) \propto q(\theta) \prod_i q(x_i | \theta)^\beta.
\end{equation}

Using a temperature to regulate the relative contribution of the prior and
posterior has been used broadly, but ordinarily doesn't have a well founded
justification. Here we can unapologetically vary the relative contributions of
the prior and likelihood since in the representational framework, those are
both variational approximations that might have differing ability to better
model the true distributions they approximate. By varying the $\beta$ parameter
here, just as in the $\beta$-VAE case~\citep{brokenelbo} we can smoothly
explore the frontier within our modeling family, smoothly controlling the
amount of information our model extracts from the dataset. This can help us
control for overfitting in a principled way.

Additionally, we could try to relax our variational approximations, and \emph{fit} our 
prior, assuming we could estimate an expectation over datasets.  One way to do that is
with a bootstrap or jackknife procedure~\citep{bootstrap}.

}%

	\section{Discriminative Models}
\label{sec:discriminative}

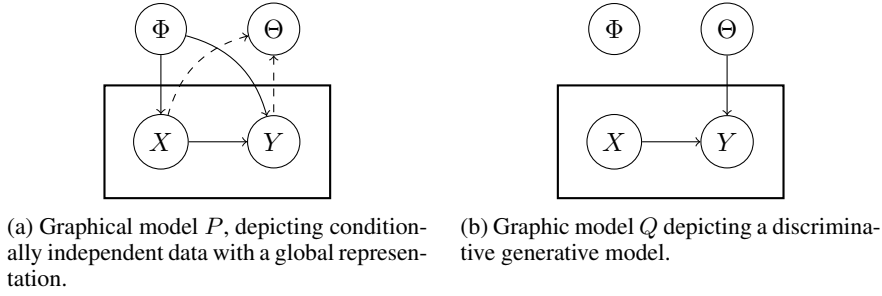
\begin{figure}[htbp]
	\centering
	\begin{subfigure}[t]{0.40\textwidth}
	\centering
	\begin{tikzpicture} 
\def\r{0.5 cm}
\node [draw, circle] (X) at (0, 0) {$X$};
\node [draw, circle] (Y) at (3*\r, 0) {$Y$};
\node [draw, circle] (theta) at (3*\r, 3*\r) {$\Theta$};
\node [draw, circle] (phi) at (0, 3*\r) {$\Phi$};
\draw [->] (X) -- (Y);
\draw [->] (phi) -- (X);
\draw [->] (phi) edge[bend left] (Y);

\draw [<-] (theta) edge[dashed]  (Y);
\draw [<-] (theta) edge[bend right, dashed] (X);

\draw [thick] (-1.5*\r, -1.5*\r) rectangle (4.5*\r, 1.5*\r);

\end{tikzpicture}
	\caption{\label{fig:dP}Graphical model $P$, depicting conditionally independent data with a 
	global representation.
	}
	\end{subfigure}
	\quad
	\begin{subfigure}[t]{0.40\textwidth}
	\centering
	\begin{tikzpicture} 
\def\r{0.5 cm}
\node [draw, circle] (X) at (0, 0) {$X$};
\node [draw, circle] (Y) at (3*\r, 0) {$Y$};
\node [draw, circle] (theta) at (3*\r, 3*\r) {$\Theta$};
\node [draw, circle] (phi) at (0, 3*\r) {$\Phi$};
\draw [->] (X) -- (Y);
\draw [->] (theta) -- (Y);

\draw [thick] (-1.5*\r, -1.5*\r) rectangle (4.5*\r, 1.5*\r);

\end{tikzpicture}
	\caption{\label{fig:dQ}Graphic model $Q$ depicting a discriminative generative model.}
	\end{subfigure}
	\caption{\label{fig:dgraphical}Graphical models for the traditional discriminative case.}
\end{figure}

Similarly we could consider the situation depicting usual discriminative
learning, depicted in \Cref{fig:dgraphical}.

For these sets of graphical models, we have the following information projection:
\begin{equation}
	\mathcal{J_\text{d}} = I_P - I_Q = I(\Theta; X^N, Y^N) + \sum_i \left[ I(X_i; \Phi) + I(Y_i; X_i, \Phi) - I(Y_i; X_i, \Theta) \right].
\end{equation}

\begin{equation}
	S \equiv \left\langle \log \frac{p(\theta|X^n)}{q(\theta)}  \right\rangle_P \geq I(\Theta; X^n)
\end{equation}
This \emph{entropy} gives an upper bound on the mutual information between our parameters and the dataset,
it requires a variational approximation to the true marginal of the posterior $p(\theta | X^N)$ over datasets: $q(\theta)$,
a \emph{prior}.

\begin{equation}
	U_i \equiv - \left\langle \log q(y_i|x_i,\theta) \right\rangle_P \geq H(Y_i | X_i, \Theta)
\end{equation}
The \emph{energy} gives an upper bound on the conditional entropy of our targets given our parameters and input, it is powered
by a variational approximation to the factored inverse of our global representation, the \emph{conditional likelihood} in ordinary 
parlance.

Our optimal frontier is set by those conditions above as well as:
\begin{equation}
	U + S \geq \mathcal{J_{\text{d}}} + \sum_i H(Y_i| X_i , \Phi) - I(X_i; \Phi)
\end{equation}

Just as previously in \Cref{sec:elwood} solutions on the frontier can be specified by:
\begin{equation}
	\log p(\theta | x^N, y^N) = \log q(\theta) + \beta \sum_i \log q(y_i|x_i,\theta) - \log \mathcal{Z} .
\end{equation}
Here again we can smoothly explore the frontier set by the variationals approximations given by the prior and
likelihood by simply adjusting $\beta$. We might additionally consider going beyond the fixed variational approximations and 
push the frontier by fitting the prior, or likelihood.

}%

	\section{Functional Inequalities}

Here we show the details for deriving \Cref{eqn:upperbound}.
\label{sec:upperboundproof}

We start by expressing our functional inequalities, but being explicit about 
presence of the relative informations of our variational approximations.
\begin{align}
	I(\Theta; X^N, Y^N) &= S - \KL{ p(\theta) }{ q(\theta) } \label{eqn:proofS} \\
	I(X_i; Z_i) &= H(X_i) - D_i + \KL{ p(x_i|z_i) }{ q(x_i | z_i) } \label{eqn:proofD}  \\
	I(Y_i; Z_i) &= H(Y_i) - C_i + \KL{ p(y_i|z_i) }{ q(y_i | z_i) } \label{eqn:proofC} \\
	I(Z_i; X_i, \Theta) &= R_i - \KL{ p(z_i) }{ q(z_i) } \label{eqn:proofR} 
\end{align}

Combining \Cref{eqn:minkl,eqn:proofS,eqn:proofD,eqn:proofC,eqn:proofR}:
\begin{equation}
	\mathcal{J} = 
	S + D + C + R - \KL{p}{q}
	- \sum_i \left[ H(X_i) + H(Y_i) - I(X_i; \Phi) - I(Y_i ; X_i, \Phi) \right] \geq 0.
\end{equation}
Here we have collected all of the KL divergences for our variational approximations:
\begin{equation}
	\begin{aligned}
		\KL{p}{q} \equiv &\KL{ p(\theta) }{ q(\theta)} + \sum_i \KL{ p(x_i|z_i) }{ q(x_i|z_i)} \\
		& + \sum_i \left[ \KL{ p(y_i|z_i) }{ q(y_i|z_i) } + \KL{ p(z_i) }{q(z_i)}\right] .
	\end{aligned}
\end{equation}

We can simplify:
\begin{align}
	H(X_i) - I(X_i; \Phi) &= H(X_i | \Phi) \\
	H(Y_i) - I(Y_i; X_i \Phi) &= H(Y_i | X_i, \Phi) \\
	H(Y_i | X_i, \Phi) + H(X_i | \Phi) &= H(Y_i , X_i| \Phi)
\end{align}
	
To obtain:
\begin{equation}
	\mathcal{J} = S + D + C + R - \KL{p}{q} - \sum_i H(Y_i, X_i | \Phi) 
\end{equation}

Which yields:
\begin{align}
	S + D + C + R &= \mathcal{J} + \KL{p}{q} + \sum_i H(Y_i, X_i | \Phi) \\
	S + D + C + R &\geq \mathcal{J} + \sum_i H(Y_i, X_i | \Phi) \\
	S + D + C + R &\geq \sum_i H(Y_i, X_i | \Phi) \\
\end{align}

%
}%

	\section{Identities}

We will utilize some basic information identities, first
by definition
\begin{align}
	I(A;B) &= H(A) - H(A|B)\\
	&= H(B) - H(B|A) \\
	&= H(A) + H(B) - H(A,B) \\
	&= H(A,B) - H(A|B) - H(B|A)
\end{align}

By the chain rule of mutual information:
\begin{equation}
	I(A,B;C) = I(A;C) + I(B;C|A) \geq 0
\end{equation}
Mutual informations, and conditional mutual informations are always positive:
\begin{align}
	I(A;B) &\geq 0 \\
	I(A;B|C) &\geq 0
\end{align}
We will also use the following rule for conditional
entropies
\begin{equation}
	H(B|A) = H(A,B) - H(A)
\end{equation}

}%

	\section{Maxwell Relations}
\label{sec:maxwell}

We can also define other potentials analogous to the 
alternative thermodynamic potentials such as enthalpy, free energy,
and Gibb's free energy
by performing partial Legendre transformations.
For instance, we can define a \emph{free rate}:
\begin{align}
	F(C,D,\sigma) &\equiv R + \sigma S \\
	dF &= -\gamma dC -\delta dD + S d\sigma.
	\label{eqn:freeenergy}
\end{align}
The free rate measures the rate of our system, not as a function of $S$
(something difficult to keep fixed), but in terms of $\sigma$, a parameter
in our loss or optimal posterior.

The free rate gives rise to other Maxwell relations such as
\begin{equation}
	\pdd{S}{C}{\sigma} = -\pdd{\gamma}{\sigma}{C},
\end{equation}
which equates how much each additional bit of entropy ($S$) buys you in terms
of classification error ($C$) at fixed effective temperature ($\sigma$), 
to a seemingly very different experiment where you measure the change in the
effective supervised tension ($\gamma$, the slope on the $R-C$ curve)
versus effective temperature ($\sigma$) at a fixed classification error ($C$).

\subsection{Complete Enumeration}
Here we enumerate a complete set of Maxwell Relations.
First if we write $R = R(D,C,S)$:
\[
	dR = -\gamma dC - \delta dD - \sigma dS 
\]

\begin{equation}
	\pdd{\gamma}{D}{C} = \pdd{\delta}{C}{D}
\end{equation}

\begin{equation}
	\pdd{\delta}{S}{D} = \pdd{\sigma}{D}{S}
\end{equation}

\begin{equation}
	\pdd{\gamma}{S}{C} = \pdd{\sigma}{C}{S}
\end{equation}

Next transforming to $F = R + \sigma S = F(D,C,\sigma)$
\[
	dF = -\gamma dC - \delta dD + S d\sigma
\]

\begin{equation}
	\pdd{\gamma}{\sigma}{C} = -\pdd{S}{C}{\sigma}
\end{equation}

\begin{equation}
	\pdd{\delta}{S}{D} = -\pdd{S}{D}{\sigma}
\end{equation}

Next transforming to $H = R + \gamma C = H(D,\gamma,S)$
\begin{equation}
	dH = C d\gamma - \delta D - \sigma dS
\end{equation}

\begin{equation}
	\pdd{C}{D}{\gamma} = -\pdd{\delta}{\gamma}{D}
\end{equation}

\begin{equation}
	\pdd{C}{S}{\gamma} = -\pdd{\sigma}{\gamma}{S}
\end{equation}

Next transforming to $G = R + \sigma S + \gamma C = G(D,\gamma,\sigma)$
\begin{equation}
	dG = C d\gamma - \delta dD + S d\sigma
\end{equation}

\begin{equation}
	\pdd{C}{\sigma}{\gamma} = \pdd{S}{\gamma}{\sigma}
\end{equation}

Next transforming to $A = R + \delta D = A(\delta, C, S)$
\begin{equation}
	dA = -\gamma dC + D d\delta - \sigma dS 
\end{equation}

\begin{equation}
	\pdd{\gamma}{\delta}{C} = -\pdd{D}{C}{\delta}
\end{equation}

\begin{equation}
	\pdd{D}{\sigma}{\delta} = -\pdd{\sigma}{\delta}{S}
\end{equation}

Finally transforming to $B = R + \delta D + \sigma S = B(\delta, C, \sigma)$
\begin{equation}
	dB = -\gamma dC + D d\delta + S d\sigma 
\end{equation}

\begin{equation}
	\pdd{\gamma}{\sigma}{C} = -\pdd{S}{C}{\sigma}
\end{equation}

\begin{equation}
	\pdd{S}{\delta}{\sigma} = \pdd{D}{\sigma}{\delta}	
\end{equation}

}%

	\section{Zeroth Law of Learning}
\label{sec:zeroth}

A central concept in thermodynamics is a notion of equilibrium. The so called
Zeroth Law of thermodynamics defines thermal equilibrium as a sort of reflexive
property of systems~\citep{finn}. If system $A$ is in thermal equilibrium with
system $C$, and system $B$ is separately in thermal equilibrium with 
system $C$, then system $A$ and $B$ are in thermal equilibrium with each other.

When any sub-part of a system is in thermal equilibrium with any other sub-part,
the system is said to be an equilibrium state.

In our framework, the points on the optimal surface are analogous to the
equilibrium states, for which we have well defined partial derivatives.
We can demonstrate that this notion of equilibrium
agrees with a more intuitive notion of
equilibrium between 
coupled systems.
Imagine we have two different models, characterized by their own
set of distributions,  Model $A$ is defined by $p_A(z|x,\theta),
p_A(\theta,\{x,y\}), q_A(z)$, and model
$B$ by $p_B(z|x,\theta), p_B(\theta,\{x,y\}),q_B(z)$.
Both models will have their own value for each of the functionals:
$R_A, S_A, D_A, C_A$ and $R_B, S_B, D_B, C_B$.  Each model defines its own 
representation $Z_A, Z_B$. Now imagine coupling the models, by forming the 
joint representation $Z_C = (Z_A, Z_B)$ formed by
concatenating the two representations together. Now
the governing distributions over $Z$ are simply the product 
of the two model's distributions, e.g.
$q_C(z_C) = q_A(z_A)q_B(z_B)$. 
Thus the rate $R_C$ and entropy $S_C$ for the combined model
is the sum of the individual models: $R_C = R_A + R_B, S_C = S_A + S_B$.

Now imagine we sample new states for the combined system
which are maximally entropic with
the constraint that the combined rate stay constant:
\begin{equation}
	\min S \text{ s.t. } R = R_C \implies p(\theta|\{x,y\})
	= \frac{q(\theta)}{\mathcal{Z}}  e^{-R/\sigma}.
\end{equation}
For the expectation of the two
rates to be unchanged after they have been coupled and 
evolved holding their total rate fixed, we must have,
\begin{equation}
	-\frac{1}{\sigma} R_A -\frac{1}{\sigma_B} R_B = - \frac{1}{\sigma_C} R_C = -\frac{1}{\sigma_C} (R_A + R_B)
	\implies \sigma_A=\sigma_B=\sigma_C.
\end{equation}
Therefore, we can see that $\sigma$, the effective temperature,
allows us to identify whether two systems are in thermal equilibrium
with one another.  Just as in thermodynamics, if two systems at
different temperatures are coupled, some transfer takes place.
}%

	\section{Experiments}
\label{sec:experiments}

We show examples of models trained on a toy dataset for all of the different
objectives we define above.  The dataset has both an infinite data variant,
where overfitting is not a problem, and a finite data variant, where
overfitting can be clearly observed for both reconstruction and classification.

\paragraph{Data generation.}
We follow the toy model from~\citet{brokenelbo}, but add an additional classification label in order to
explore supervised and semi-supervised objectives.
The true data generating distribution is as follows.  We first sample a latent
binary variable, $z \sim \Ber(0.7)$, then sample a latent 1D continuous value
from that variable, $h|z \sim \gauss(h|\mu_z,\sigma_z)$, and finally we observe
a discretized value, $x = \mathrm{discretize}(h;\calB)$, where $\calB$ is a set
of 30 equally spaced bins, and a discrete label, $y = z$ (so the true label is the latent variable that generated $x$).
We set $\mu_z$ and $\sigma_z$ such that $R^* \equiv I(x;z)=0.5$ nats, in the
true generative process, representing the ideal rate target for a latent
variable model.  For the finite dataset, we select 50 examples randomly from
the joint $p(x,y,z)$.  For the infinite dataset, we directly supply the true
full marginal $p(x,y)$ at each iteration during training.  When training on the
finite dataset, we evaluate model performance against the infinite dataset so
that there is no error in the evaluation metrics due to a finite test set.

\paragraph{Model details.}
We choose to use a discrete latent representation with $K=30$ values, with an
encoder of the form $q(z_i|x_j) \propto -\exp[(w^e_i x_j - b^e_i)^2]$, where
$z$ is the one-hot encoding of the latent categorical variable, and $x$ is the
one-hot encoding of the observed categorical variable.
We use a decoder of the same form, but with different
parameters: $q(x_j|z_i) \propto -\exp[(w^d_i x_j - b^d_i)^2]$.
We use a classifier of the same form as well: $q(y_j|z_i) \propto -\exp[(w^c_i y_j - b^c_i)^2]$.
Finally, we use a variational marginal, $q(z_i)=\pi_i$.
Given this, the true joint distribution has the form
$p(x,y,z) = p(x) p(z|x) p(y|x)$, with marginal
$p(z) = \sum_x p(x,z)$, and conditionals
$p(x|z) = p(x,z)/p(z)$ and $p(y|z) = p(y,z)/p(z)$.

The encoder is additionally parameterized following~\citet{emergence} by
$\alpha$, a set of learned parameters for a Log Normal distribution of the form
$\log \calN(-\alpha_i/2, \alpha_i)$. In total, the model has 184 parameters: 60
weights and biases in the encoder and decoder, 4 weights and biases in the
classifier, 30 weights in the marginal, and an additional 30 weights for the
$\alpha_i$ parameterizing the stochastic encoder. We initialize the weights so
that when $\sigma=0$, there is no noticeable effect on the encoder during
training or testing.

\paragraph{Experiments.}
In Figure~\ref{fig:toymodelopt}, we show the optimal,
hand-crafted model for the toy dataset, as well as a selection of
parameterizations of the TherML objective that correspond to commonly-used
objective functions and a few new objective functions not previously described.
In the captions, the parameters are specified with $\gamma, \delta, \sigma$ as
in the main text, as well as $\rho$, which is a corresponding Lagrange
multiplier for $R$, in order to simplify the parameterization.  
It just
parameterizes the optimal surface slightly differently.  We train all
objectives for 10,000 gradient steps. For all of the objectives described, the
model has converged, or come close to convergence, by that point.

Because the model is sufficiently powerful to memorize the dataset, most of the
objectives are very susceptible to overfitting. Only the objective variants
that are ``regularized'' by the $S$ term (parameterized by $\sigma$) are able
to avoid overfitting in the decoder and classifier.

\def \supwidth {0.45}

\begin{figure}[htbp]
    \centering
    \includegraphics[width=\supwidth\textwidth]{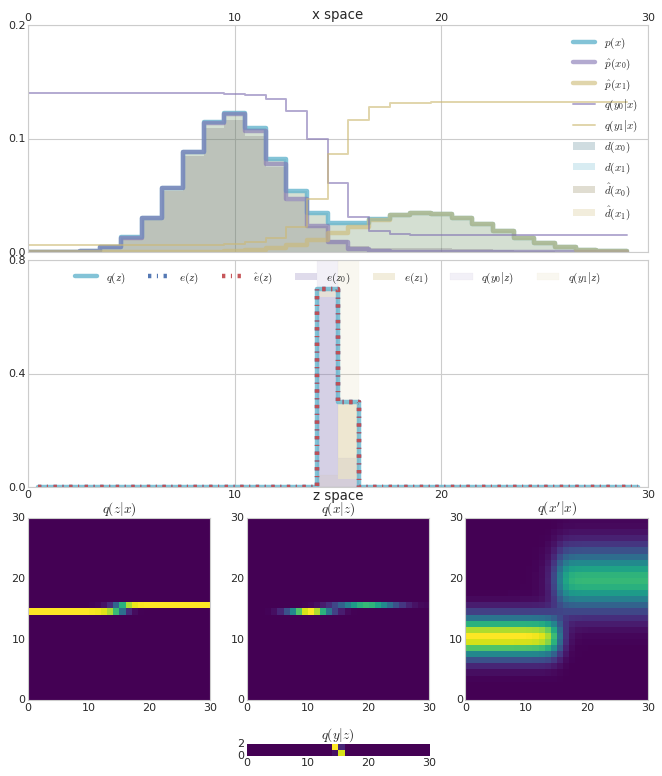}
	\caption{\label{fig:toymodelopt}\textbf{Hand-crafted optimal model.}
     \small
     Toy Model illustrating the difference between selected points on the three dimensional optimal surface defined by $\gamma$, $\delta$, and $\sigma$.
     See Section~\ref{sec:frontier} for more description of the objectives, and Appendix~\ref{sec:experiments} for details on the experiment setup.
     {\bf Top (i):} Three distributions in data space: the true data
           distribution, $p(x)$, the model's generative distribution, $g(x) = \sum_z
           q(z) q(x|z)$, and the empirical data reconstruction distribution, $d(x) =
           \sum_{x'} \sum_z p(x') q(z|x') q(x|z)$.
     {\bf Middle (ii):} Four distributions in latent space: the learned (or
           computed) marginal $q(z)$, the empirical induced marginal $e(z) = \sum_x
           p(x) q(z|x)$, the empirical distribution over $z$ values for data vectors
           in the set $\calX_0 = \{ x_n: z_n = 0\}$, which we denote by $e(z_0)$ in
           purple, and the empirical distribution over $z$ values for data vectors in the
           set $\calX_1 = \{ x_n: z_n = 1\}$, which we denote by $e(z_1)$ in yellow.
     {\bf Bottom:} Three $K \times K$ distributions: (iii) $q(z|x)$, (iv) $q(x|z)$ and
     (v) $q(x'|x)=\sum_z q(z|x) q(x'|z)$.
	}
\end{figure}


\begin{figure*}
  \begin{subfigure}{\supwidth\textwidth}
    \includegraphics[width=1.0\textwidth]{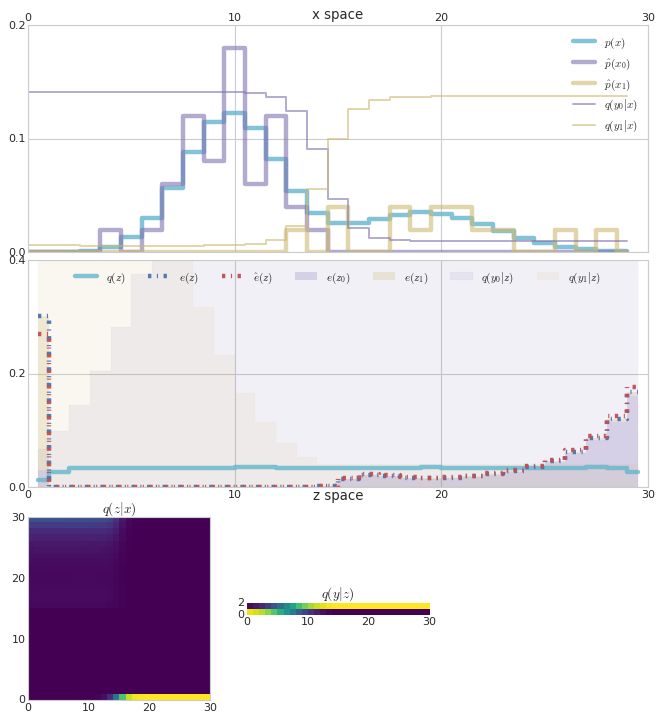}
    \caption{\label{fig:toymodeldetermclass}\textbf{Deterministic Supervised Classifier:} $\delta = \rho = \sigma = 0, \gamma = 1$.}
  \end{subfigure}
	\hfill
  \begin{subfigure}{\supwidth\textwidth}
    \includegraphics[width=1.0\textwidth]{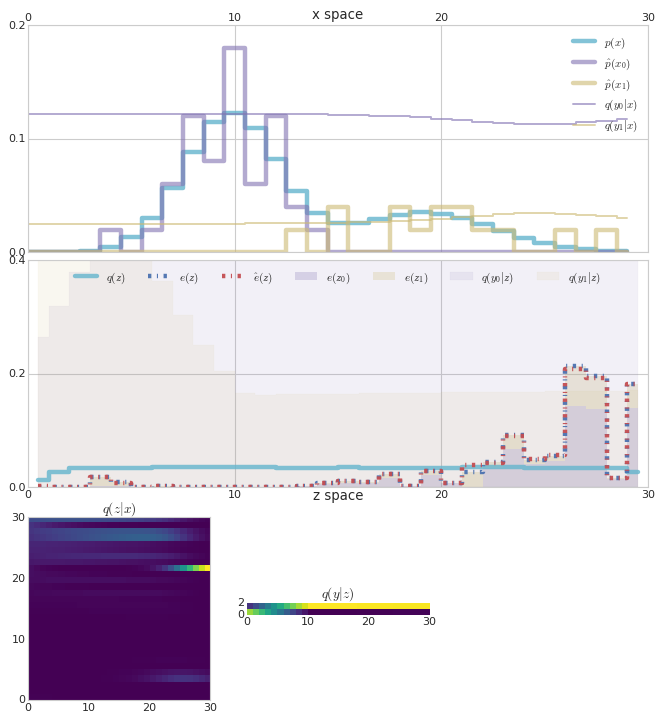}
    \caption{\label{fig:toymodelentregdeterm}\textbf{Entropy-regularized Deterministic Classifier:} $\delta = \rho = 0, \gamma = 1, \sigma = 0.1$.}
  \end{subfigure}

  \begin{subfigure}{\supwidth\textwidth}
    \includegraphics[width=1.0\textwidth]{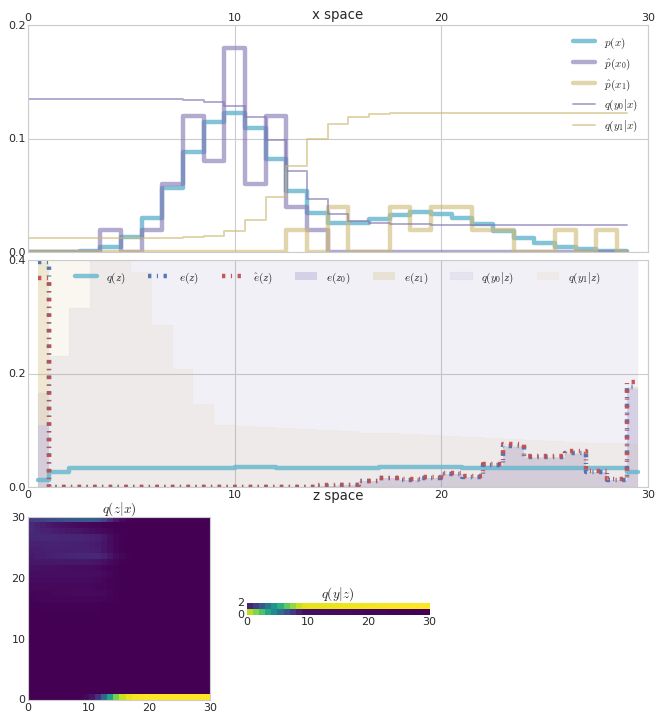}
    \caption{\label{fig:toymodelentib}\textbf{Entropy-regularized IB:} $\delta = 0, \rho = 0, \gamma = 1, \sigma = 0.01$.}
  \end{subfigure}
\hfill
  \begin{subfigure}{\supwidth\textwidth}
    \includegraphics[width=1.0\textwidth]{figures/toy/therml_rrg50g5_3_5_kC_1_kS_1_klr_0_001_opt_adam_train_op_erpow_1_drpow_1_report.png}
    \caption{\label{fig:toymodelbbb}\textbf{Bayesian Neural Network Classifier:} $\delta = 0, \rho = 0, \sigma = \gamma = 1$.}
  \end{subfigure}
  \caption{\label{fig:supervised}Supervised Learning approaches.}
\end{figure*}

\def \unsupwidth {\supwidth}
\begin{figure*}
  \begin{subfigure}{\unsupwidth\textwidth}
    \includegraphics[width=1.0\textwidth]{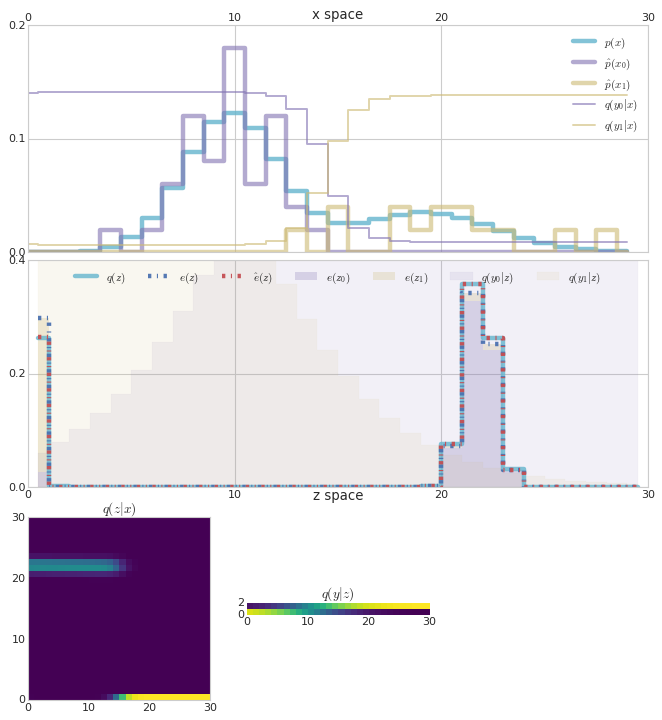}
    \caption{\label{fig:toymodelvib}\textbf{VIB:} $\delta = 0, \sigma = 0, \gamma = 1, \rho (\beta) = 0.5$.}
  \end{subfigure}
	\hfill
  \begin{subfigure}{\unsupwidth\textwidth}
    \includegraphics[width=1.0\textwidth]{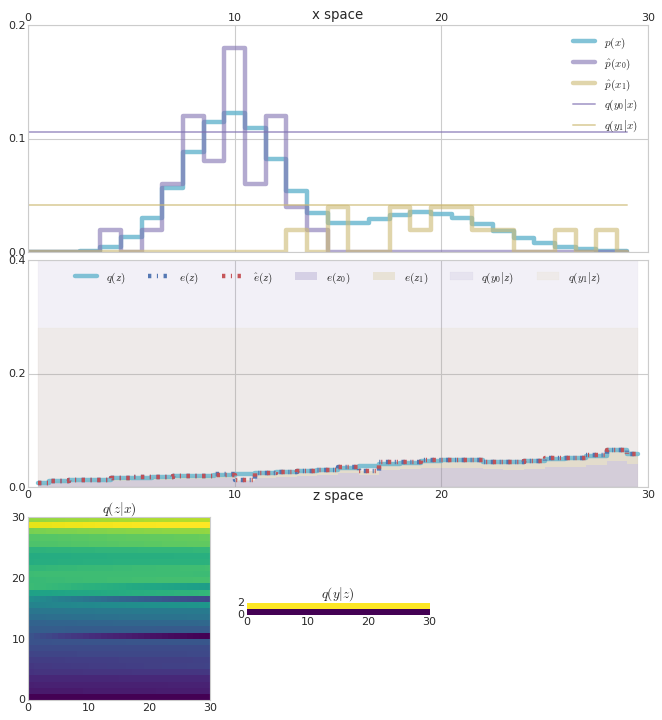}
    \caption{\label{fig:toymodelentvib}\textbf{Entropy-regularized VIB:} $\delta = 0, \gamma = 1, \rho = 0.9, \sigma = 0.1$.}
  \end{subfigure}
	\caption{\label{fig:vib}VIB style objectives.}
\end{figure*}

\begin{figure*}
  \begin{subfigure}{\unsupwidth\textwidth}
    \includegraphics[width=1.0\textwidth]{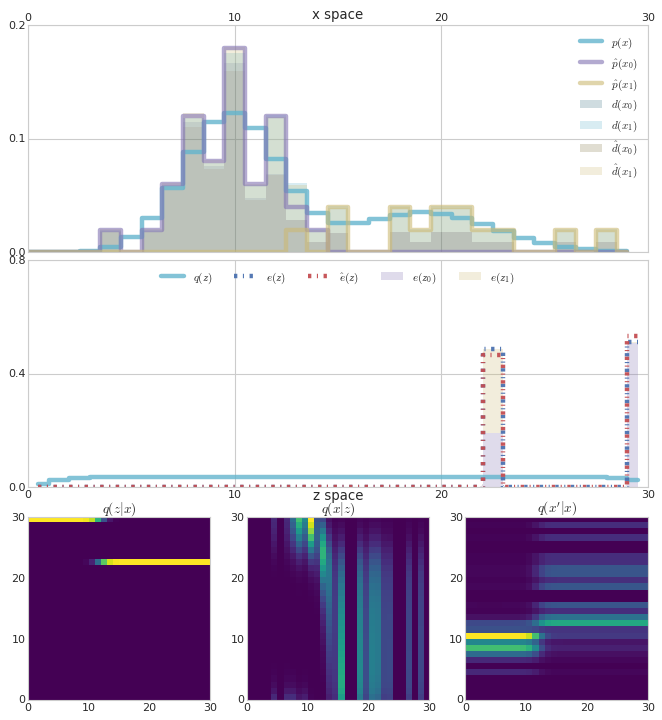}
    \caption{\label{fig:toymodeldetermautoencoder}\textbf{Deterministic Autoencoder:} $\gamma = \rho = \sigma = 0, \delta = 1$.}
  \end{subfigure}
	\hfill
  \begin{subfigure}{\unsupwidth\textwidth}
    \includegraphics[width=1.0\textwidth]{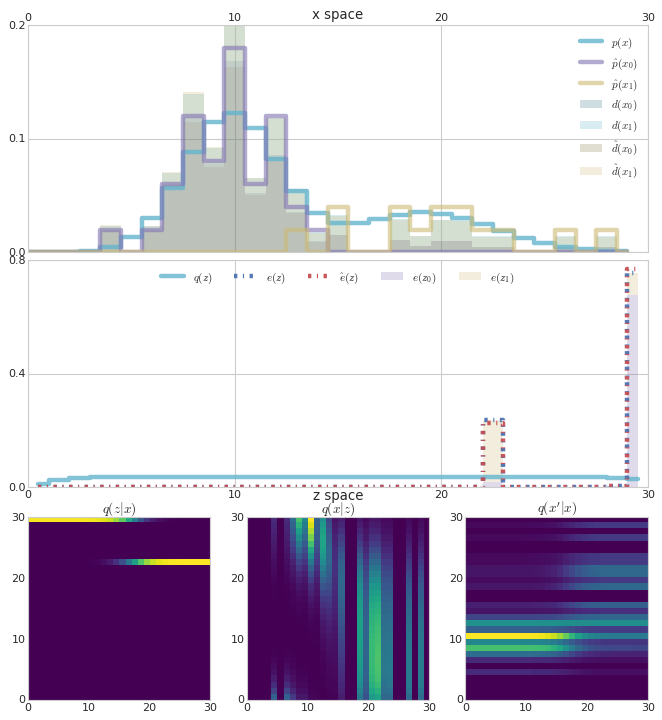}
    \caption{\label{fig:toymodelentautoencoder}\textbf{Entropy-regularized Deterministic Autoencoder:} $\gamma = \rho = 0, \delta = 1, \sigma = 0.01$.}
  \end{subfigure}
	\caption{\label{fig:ae}Autoencoder objectives.}
\end{figure*}

\def \vaewidth {\supwidth}
\begin{figure*}
	\centering
  \begin{subfigure}{\unsupwidth\textwidth}
    \includegraphics[width=1.0\textwidth]{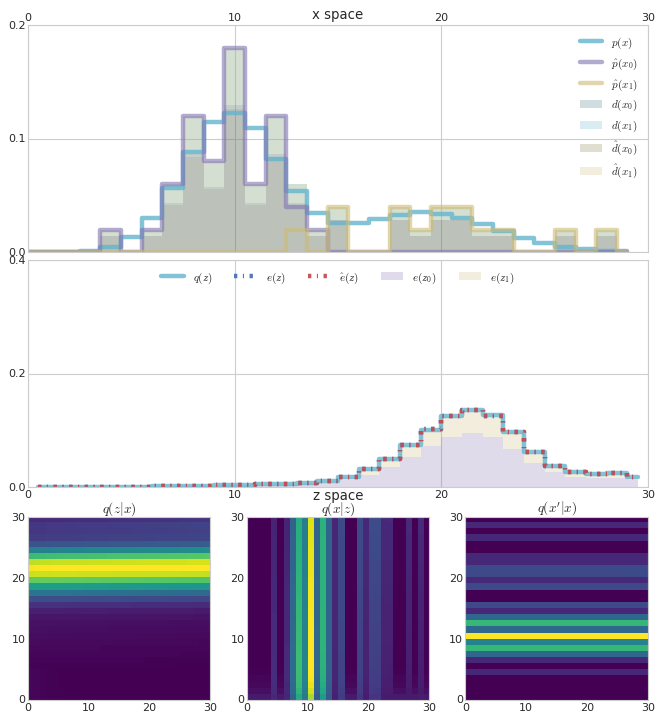}
    \caption{\label{fig:toymodelelbo}\textbf{VAE:} $\sigma = 0, \gamma = 0, \delta = \rho = 1$.}
  \end{subfigure}
\hfill
  \begin{subfigure}{\unsupwidth\textwidth}
    \includegraphics[width=1.0\textwidth]{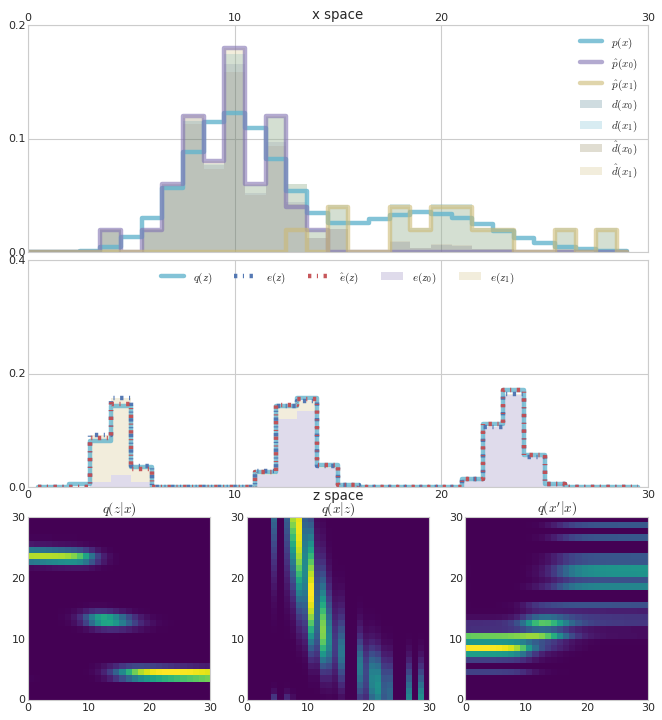}
    \caption{\label{fig:toymodelbetavae}\textbf{$\beta$-VAE:} $\sigma = 0, \gamma = 0, \delta = 1, \rho (\beta) = 0.5$.}
  \end{subfigure}

  \begin{subfigure}{\unsupwidth\textwidth}
    \includegraphics[width=1.0\textwidth]{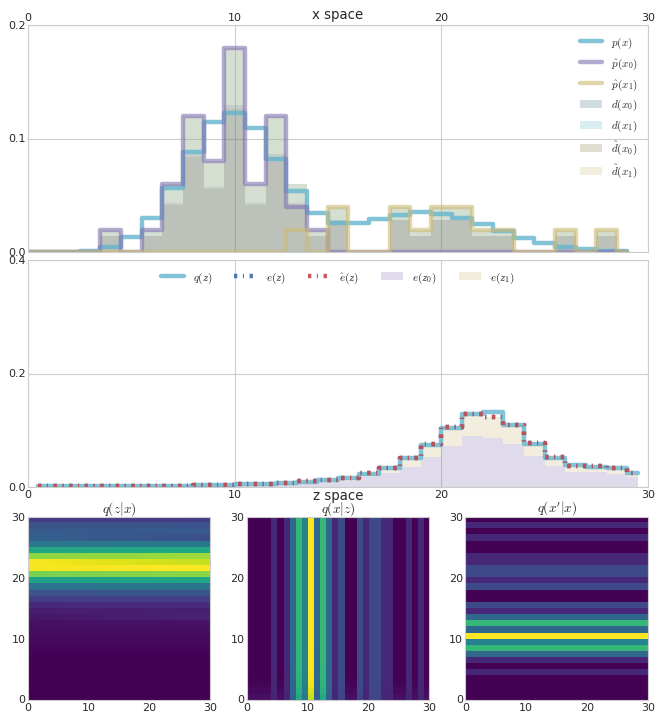}
    \caption{\label{fig:toymodelentbetavae}\textbf{Entropy-regularized $\beta$-VAE:} $\sigma = 0.5, \gamma = 0, \delta = 1, \rho (\beta) = 0.9$.}
  \end{subfigure}
\hfill
  \begin{subfigure}{\unsupwidth\textwidth}
    \includegraphics[width=1.0\textwidth]{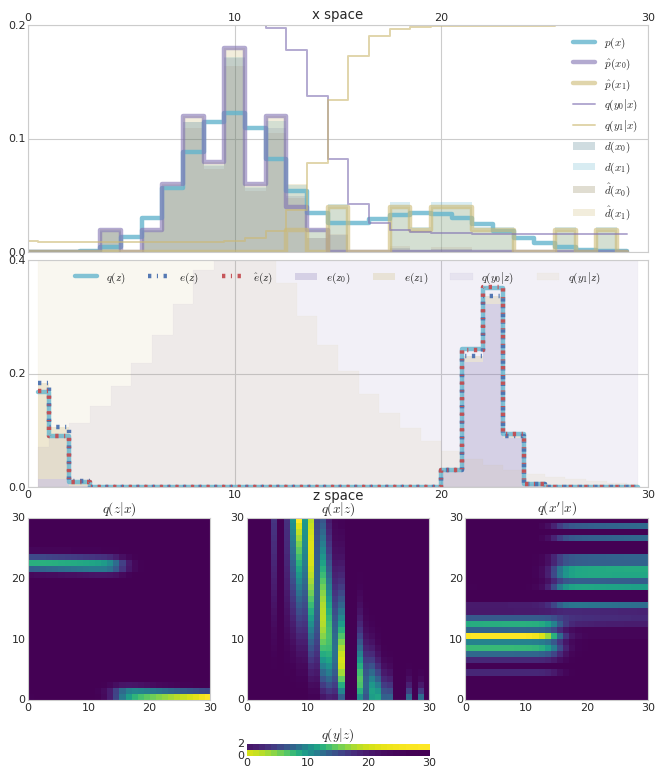}
    \caption{\label{fig:toymodelsemivae}\textbf{Semi-supervised VAE:} $\sigma = 0, \gamma (\alpha) = 0.5, \delta = \rho = 1$.}
  \end{subfigure}
	\caption{\label{fig:vae}VAE style objectives.}
\end{figure*}

\begin{figure*}[htbp]
    \centering
    \includegraphics[width=\supwidth\textwidth]{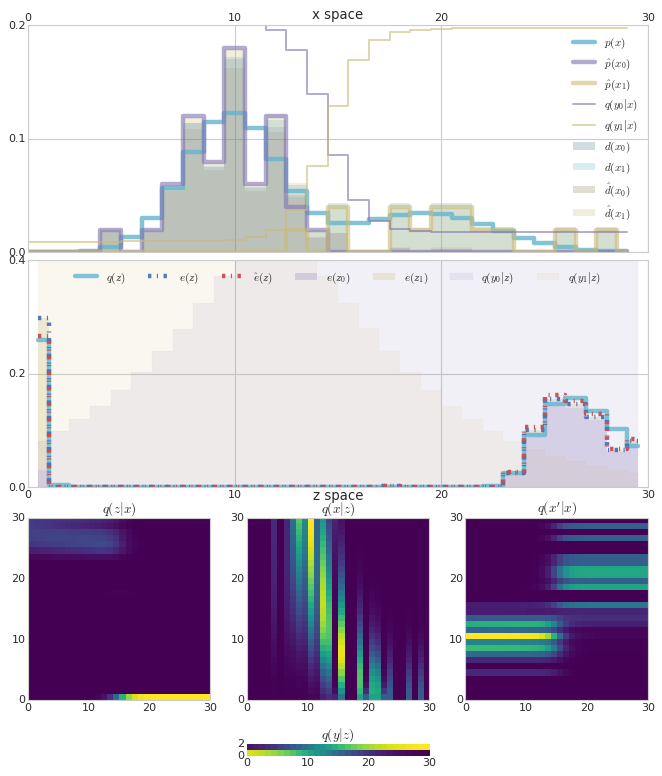}
	\caption{\label{fig:fullguy}\textbf{Full Objective.} $\sigma = 0.5, \gamma = 1000, \delta = 1, \rho = 0.9$.
     \small
	Simple demonstration of the behavior with all terms present in the objective.
	}
\end{figure*}

}%


\end{document}